\documentclass{article}
\usepackage{spconf,amsmath,graphicx}
\usepackage{amsthm,amsmath,amssymb}
\usepackage{mathrsfs}
\usepackage{fdsymbol}
\usepackage{booktabs}
\usepackage{multirow}
\usepackage{url}

\usepackage{xcolor}

\title{Towards Efficiently Diversifying Dialogue Generation \\ via Embedding Augmentation}
\name{Yu Cao$^{\dagger}$, Liang Ding$^{\dagger}$\sthanks{This work was done when Yu Cao and Liang Ding were interning at Tencent AI Lab. Code is available at:  https://github.com/caoyu-noob/embedding\_augmentation }, Zhiliang Tian$^{\mathsection}$, and Meng Fang$^{\ddagger}$}
\address{$^{\dagger}$ School of Computer Science, The University of Sydney, Australia, $^{\ddagger}$ Tencent Robotics X, China \\ 
$^{\mathsection}$ Department of CSE, The Hong Kong University of Science and Technology, Hong Kong\\
}
\begin{document}
%
\maketitle
\begin{abstract}
Dialogue generation models face the challenge of producing generic and repetitive responses. Unlike previous augmentation methods that mostly focus on token manipulation and ignore the essential variety within a single sample using hard labels, we propose to promote the generation diversity of the neural dialogue models via soft embedding augmentation along with soft labels in this paper. Particularly, we select some key input tokens and fuse their embeddings together with embeddings from their semantic-neighbor tokens. The new embeddings serve as the input of the model to replace the original one. Besides, soft labels are used in loss calculation, resulting in multi-target supervision for a given input. Our experimental results on two datasets illustrate that our proposed method is capable of generating more diverse responses than raw models while remains a similar n-gram accuracy that ensures the quality of generated responses.
\end{abstract}
\section{Introduction}
\label{sec:intro}

Dialogue generation is important in AI applications, in which a model can generate responses for user-issued dialogue history. Former dialogue datasets make the end-to-end training of deep neural models possible, such as DailyDialog~\cite{dailydialog}, PersonaChat~\cite{personachat}, etc., and deep neural networks including Seq2seq with attention~\cite{seq2seq}, Transformers~\cite{transformer} have already shown their capability in generating conversation replies. Dialogue generation can be regarded as a \textbf{many-to-many} (combination of one-to-many and many-to-one) problem as one specific response may be reasonable for multiple histories and vice versa. However, current models tend to produce generic sentences such as ``I don't know'', caused by the current inherent deterministic training objective as well as insufficient diversity and limited quality of current datasets~\cite{diversity_objective,shao_quality}. 

Generating more various dialogue responses remains a hard task. In order to tackle it, some researchers try to refine the training objective with extra constraint items~\cite{diversity_objective} or modify the criterion that encourages models to decode more diverse words~\cite{kurata_generate,kulikov_decoding,li2016simple,song_diversifying}. However, such models are more likely to generate ungrammatical or uncorrelated responses and the superiority of constructed objectives to cross-entropy loss remains unknown. 
Another kind of approach enhances the training data directly. Data filtering is commonly used in machine learning and extended to dialogue generation by removing samples with generic responses based on entropy~\cite{entropy_filter} or the predictions of a Seq2seq model~\cite{data_distillation}. 
Obtaining more training samples is also a possible solution. Word replacement extends the data scales by randomly replacing original tokens with others based on vocabulary distributions~\cite{xie_replacement,fadaee_replacement}. 
In addition, new augmented sentences can also be obtained via a learnable model that even keeps interaction between the generation model and both of them will be trained jointly~\cite{insufficient_rock}. But these methods merely include more training data and still bridge a one-to-one mapping between input and response within a single sample. And they also add great extra computational load. So it is necessary to propose a simple and effective method regarding the essence of this problem.

In this paper,  we propose an embedding augmentation method for dialogue models, where not only the training objective encourages various output, but also soft embeddings enhance the diversity within a single sample. It is inspired by the recent success of the mixup approach that combines training pairs of samples and labels convexly into a single one~\cite{mixup}. 
Particularly, the original embeddings of tokens in the training samples will be randomly replaced by augmented ones, which are mixtures of the raw one and several semantically similar embeddings conditioned on their distribution. And these selected positions will use soft labels instead of hard one-hot vectors. Such a mechanism ensures that the model can learn a \textit{soft word distribution} rather than a fixed word in both input and output, which is consistent with the multi-source and multi-target purpose, benefiting the generating diversity without using any additional sample. Compared to previous similar work~\cite{kobayashi_contextual,gao_mixup}, we use both soft embedding and soft labels for a more flexible training process. A compact in-domain similar token prediction model is utilized instead of a deep language model, realizing a more efficient augmentation with a less computational cost.

To verify our method, we conduct experiments on two dialogue datasets, PersonaChat and DailyDialog, using two base models, Seq2seq and transformer. The experimental results show that our method can obtain remarkable diversity improvement as well as n-gram accuracy boost in terms of auto metrics compared to models without embedding augmentation or using token-level replacement. We also compare it with a BERT-based embedding augmentation method, demonstrating they have close performance while the former one is much faster and proving the efficiency of our method.

\section{Method}


\subsection{Background and motivations}

Generally speaking, given a dialogue history $H=(h_1, h_2,...,$ $h_N)$ and a corresponding response $R=(r_1,r_2,...,r_M)$, current neural dialogue generation models learn the conditional probability $p(r_1,r_2,...,r_M|h_1,h_2,...,h_N)$ which is similar to machine translation. 
But as discussed before, dialogue tasks should be many-to-many that differs from machine translation who has higher certainty. Nevertheless, there is usually one specific $R$ that corresponds to a given $H$ in most datasets. Although one-to-many or many-to-one cases exist, most of them are generic sentences having no contribution to the variety of generated replies~\cite{entropy_filter}. Thus existing samples may provide the wrong guidance for the task goal.

Some existing augmentation methods generate new samples by replacing tokens in original sentences with others~\cite{fadaee_replacement,kobayashi_contextual}. But such a hard replacement will also magnify the error if inappropriate candidate tokens are introduced. Although soft word distribution is employed in ~\cite{gao_mixup}, it still uses hard labels for loss calculation. Such augmentation only ensures a many-to-one learning objective for the model, differing from our desired target. These motivate us to use soft word embeddings and soft labels in augmentation. The use of fused embedding can lower the influence of unbefitting tokens as partial original information still remains in the embedding. While the latter one can bring randomness and multiple targets during training to further diversify the generation.

\subsection{Embedding augmentation}

\begin{figure}[ht!]

\begin{minipage}[b]{1.0\linewidth}
  \centering
  \includegraphics[width=8.0cm]{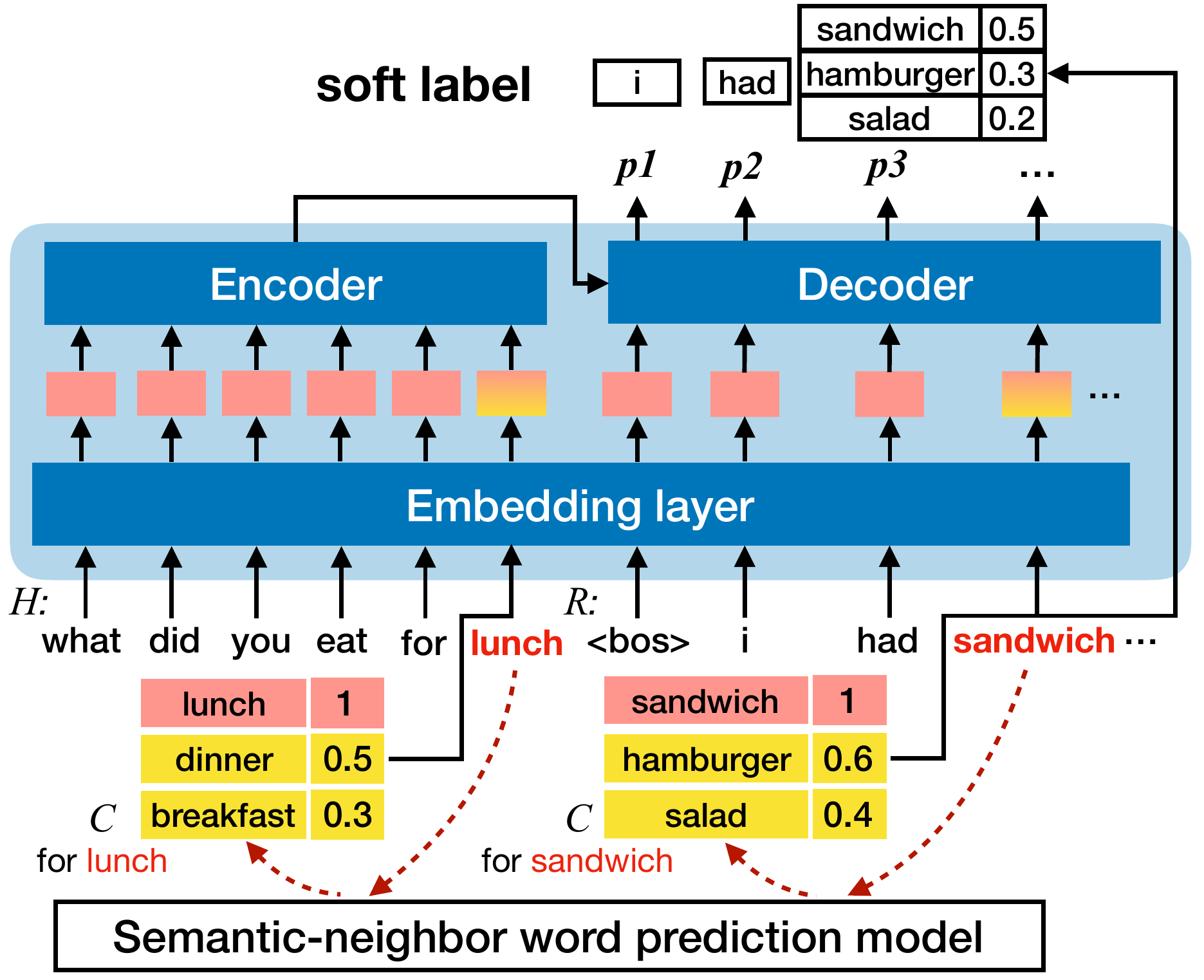}
\end{minipage}
\caption{Framework of our embedding augmentation method.}
\label{fig:framework}
\end{figure}

Inspired by the above intuition, we provide an embedding augmentation method, whose framework is shown in Fig.~\ref{fig:framework}. An encoder-decode model is used for generation. The history $H$ is encoded to get its hidden state as the initial state of decoder. The model is trained with supervision from the response $R$.
Embeddings of selected tokens in $H$ and $R$ will be replaced with augmented ones from distributions of their semantic neighbors. These distributions are also used as labels.

We first select target tokens for augmentation with some exceptions, which is indicated by the red font in Fig.~\ref{fig:framework}. Since special tokens including {\bf commas}, {\bf articles} (a/an/the) and {\bf prepositions} (in/on, etc.) are usually unique, replacing their embedding has slight benefit for diversity or even causes semantically degeneration. They are not considered as target tokens. The rest tokens are randomly selected with augmentation ratio $\rho$, resulting in a set target set $\mathcal{T}=(t_1,...,t_{n+m})$ with $n$ tokens from $H$ and $m$ tokens from $R$.

Then a semantic-neighbor word prediction model is utilized to predict the candidate words with the most similar meaning for each token in $\mathcal{T}$. It prevents semantic ambiguity when mixing the embeddings from these candidates together. A most intuitive method is using a trained large-scale language model such as BERT~\cite{bert} to contextually predict candidates. But considering the computational efficiency of our method, we use \textit{fastText} model here~\cite{fasttext}, and the effectiveness of using this model will be later illustrated in the experiment. It is based on a continuous-bag-of-word frame which is fast to train. 
Given a word, it will output a series of candidates $c_i \in \mathcal{V}$ with confidence scores $s_i \in (0,1)$.
We will train the fastText model using the same corpus as the training data for the dialogue model, so as to get a consistent vocabulary $\mathcal{V}$ as well as in-domain word distribution.
To ensure the quality of augmentation, only candidates with top-\textit{k} confidence remain, and their scores must satisfy $s_i \geq \tau$ where $\tau$ is a threshold. 

One target token $t$ and its semantic neighbors can form a soft word set $\mathcal{C}=\{(c_0,s_0),(c_1,s_1),...,$ $(c_k,s_k)\}, c_0=t$, here $s_0=1$ keeps a higher proportion of the raw information as we believe it is more reliable than predicted candidates. $c_j$ in this set can then be transformed into a set of new representations $\{e_0,e_1,...,e_k\}, e_j \in \mathbb{R}^d$ via the embedding layer. The fused embedding $e^f$ of them for $t$ can be obtained via
\vspace{-0.5mm}
\begin{equation}
    e^f=\sum{_{j=0}^{j=k} \ {p(c_j) \cdot e_j}}, {\rm{where }} p(c_j)=s_j / \sum{_{l=0}^{l=k} \ {s_l}}.
    \label{eq:fuse}
\end{equation}
It can be expected as an augmented soft embedding based on the distribution of the original token and its semantic neighbors to replace the raw embedding. The distribution $p(c_j)$ is calculated using scores from the neighbor prediction model.

It is worth pointing out that no external knowledge is included as we just depend on the original dataset, which is different from some previous works~\cite{cai_reweight,kobayashi_contextual}. And our method can be easily extended to other different base models as the only model-level modification is the embedding layer.

\subsection{Model training}

The cross-entropy loss will be employed as the optimization objective. Given the generation logits $g \in \mathbb{R}^{|\mathcal{V}|}$ after softmax function for current position $i$ in label (response), if $r_i$ is selected as the target augmentation token, the loss for $r_i$ is
\vspace{-0.5mm}
\begin{equation}
    L_i=-\sum{_{j=0}^{j=k}} \ {p(c_{ij}) \cdot log \ g(c_{ij})},
\end{equation}
where $c_{ij}$ is a token from the candidate set $\mathcal{C}$ for $r_i$, and $p(c_j)$ is the distribution among candidates as (\ref{eq:fuse}), $g(c_j)$ denotes getting the corresponding value in vector $g$ for $c_j$. If $r_i$ is not an augmentation token, the training loss becomes $L_i=-log \ g(r_i)$, which is the same as the previous models.

Moreover, the augmentation operation is executed alternately along with the normal training procedure without augmentation in each step. This aims to guarantee a correct learning direction by original samples and prevent possible continuous error propagation from augmented embeddings.

\section{Experiment}

\subsection{Settings}

Two public dialogue datasets, PersonaChat~\cite{personachat} and DailyDialog~\cite{dailydialog} are involved in our experiment, containing 131k/ 7.8k/ 7.5k and 42.6k/ 3.9k/ 3.7k for training/validation/test set respectively. The former one contains persona profiles as auxiliary information while the latter one contains topic information. We will concatenate it with the original dialog history as the input $H$, using a special token to distinguish each part.

We consider two base end-to-end models, Seq2Seq with Attention (\textbf{Seq2seq})~\cite{seq2seq} using 1-layer bidirectional GRU as the encoder and decoder, and transformer model (\textbf{transfo})~\cite{transformer} in which the encoder and decoder both have 6 layers and 4 heads in attention. 300d GloVe~\cite{glove} is used to initiate the embedding layer. The hidden dimension for both models is set as 300. Each model on each dataset is trained via Adam optimizer with batch size 256 for 50 epochs(PerosnaChat)/ 30 epochs(DailyDialog). During decoding, beam search with size 3 is applied to the model with the best PPL. Besides, a fastText model will be trained on each dataset for 100 epochs using the `cbow' method (a config of fastText means continuous bag-of-word), then used as the neighbor prediction model correspondingly. In augmentation, the augmentation ratio $\rho$ is 0.4 for PersonaChat and 0.5 for DailyDialog, the score threshold $\tau$ is 0.4, and top-ranking number $k$ is 5.

Except for our embedding augmentation (\textbf{-EA}), we also consider two baselines: 1) the original model without augmentation; 2) a replacement baseline (\textbf{-rep}) in which a token is directly replaced by the most similar one in its neighbor candidates. Besides, one variant is included that uses the original BERT to replace fastText (\textbf{-BERT}) for augmentation candidates prediction while the rest parts remain the same. 

We use multiple automatic metrics to evaluate the performance. \textbf{BLEU}~\cite{bleu}, \textbf{METEOR}~\cite{meteor} and \textbf{NIST-4}~\cite{tian_response} are used for n-gram accuracy of generated replies. Following metrics are considered for measuring diversity, 1) \textbf{Ent-n}~\cite{entropy} calculate the entropy based on the appearance probabilities of each n-gram in all predictions; 2) \textbf{Dist-n} (distinct)~\cite{diversity_objective} is defined as the ratio of unique n-grams over all n-grams in all generated sentences; 3) \textbf{Sen-n} is the average of sentence-level \textbf{Dist-n}(n=1,2,3). Besides, generation perplexity (PPL) and average length (avg.len) of responses are also provided.

\subsection{Main results}

\begin{table*}[t!]
\setlength{\tabcolsep}{1mm}
\centering
\begin{tabular}{c|l|cc|ccc|ccccccccc}
\hline 
& \textbf{Model} & \small PPL & \small avg.len & \small BLEU & \small MET & \small N-4 & Ent-1 & Ent-2 & Ent-3 & Dist-1 & Dist-2 & Dist-3 & Sen-1 & Sen-2 & Sen-3 \\
\hline
\multirow{6}*{\rotatebox{90}{PersonaChat}} & Seq2seq & 36.016 & 8.462 & 2.508 & 7.636 & 1.003 & 4.114 & 5.992 & 7.326 & 1.811 & 8.646 & 20.286 & \bf 89.547 & 96.609 & 98.211  \\
~ & Seq2seq-rep & 36.814 & 8.579 & 2.685 & 8.135 & 1.168 & \bf 4.156 & 6.115 & 7.509 & \bf 1.888 & 9.227 & 21.032 & 87.628 & 96.079 & 98.756 \\
~ & \textbf{Seq2seq-EA} & \bf 35.368 & 8.651 & \bf 2.729 & \bf 8.206 & \bf 1.274 & 4.137 & \bf 6.159 & \bf 7.564 & 1.802 & \bf 9.653 & \bf 21.891 & 88.547 & \bf 96.684 & \bf 98.847\\
\cline{2-16}
~ & transfo & 39.882 & 8.602 & 2.839 & 7.840 & 1.107 & 4.112 & 5.570 & 6.313 & 1.532 & 5.358 & 10.059 & \bf 92.088 & 97.432 & 98.390 \\
~ & transfo-rep & 35.993 & 8.620 & 3.031 & 8.121 & 1.175 & 4.240 & 5.844 & 6.696 & \bf 1.771 & 6.787 & 12.733 & 90.793 & 96.963 & 98.227 \\
~ & \textbf{transfo-EA} & \bf 34.480 & 8.652 & \bf 3.450 & \bf 8.414 & \bf 1.204 & \bf 4.269 & \bf 5.963 & \bf 6.821 & 1.753 &  \bf 6.809 & \bf 12.927 & 91.387 & \bf 97.833 & \bf 98.756 \\
\hline
\multirow{6}*{\rotatebox{90}{DailyDialog}} & Seq2seq & 53.707 & 7.008 & 1.108 & \bf 5.471 & \bf 0.291 & \bf 4.515 & 6.579 & \bf 7.691 & 4.956 & 18.038 & \bf 35.422 & 92.098 & 96.953 & \bf 92.725 \\
~ & Seq2seq-rep & 54.944 & 6.560 & 1.008 & 5.140 & 0.162 & 4.436 & 6.411 & 7.425 & 4.980 & 18.524 & 33.857 & \bf 94.923 & \bf 98.116 & 88.474 \\
~ & \textbf{Seq2seq-EA} & \bf 53.541 & 6.804 & \bf 1.214 & 5.452 & 0.217 & 4.455 & \bf 6.648 & 7.657 & \bf 5.095 & \bf 19.459 & 34.543 & 94.359 & 97.941 & 92.691 \\
\cline{2-16}
~ & transfo & 77.952 & 7.009 & 1.005 & 5.032 & 0.202 & 3.923 & 5.375 & 6.087 & 2.296 & 7.692 & 12.925 & 94.076 & 97.877 & 87.817 \\
~ & transfo-rep & 74.989 & 6.732 & 1.051 & 5.025 & \bf 0.207 & \bf 4.089 & 5.576 & \bf 6.374 & \bf 3.033 & \bf 9.434 & 16.090 & 94.820 & 98.171 & 89.431 \\
~ & \textbf{transfo-EA} & \bf 74.020 & 6.835 & \bf 1.097 & \bf 5.106 & 0.198 & 4.052 & \bf 5.631 & 6.365 & 2.989 & 9.263 & \bf 16.658 & \bf 95.280 & \bf 98.187 & \bf 90.608 \\
\hline
\end{tabular}
\caption{\label{tab:results}Automatic evaluation results on two datasets (\% for BLEU, Dist-n and Sen-n). MET: METEOR, N-4: NIST-4.}
\end{table*}

\begin{table}[t!]
\setlength{\tabcolsep}{0.9mm}
\small
\centering
\begin{tabular}{l|c|cc|ccc}
\hline 
\textbf{Model} & $\Delta$PPL & $\Delta$BLEU & $\Delta$N-4 & $\Delta$Ent & $\Delta$Dist & $\Delta$Sen  \\
\hline
Seq2seq-BERT & \textit{1.212} & 0.050 & \textit{-0.012} & 0.032 & \textit{-0.130} & \textit{-0.738} \\
\hline
transfo-BERT & 1.688 & \textit{-0.108} & \textit{-0.023} & \textit{-0.116} & \textit{-0.091} & \textit{-0.743} \\
\hline
\hline
Seq2seq-BERT & -1.885 & \textit{-0.155} & \textit{-0.082} & \textit{-0.046} & 0.597 & \textit{-0.408} \\
\hline
transfo-BERT & \textit{2.860} & \textit{-0.011} & 0.071 & 0.104 & 1.057 & \textit{-0.739} \\
\hline
\end{tabular}
\caption{\label{tab:bert}Performance value difference ($\Delta$) when the BERT variant of our method compared to our original one. Upper: PersonaChat, lower:DailyDialog. (Italic: -BERT is worse)}
\end{table}

The augmentation results along with two baselines, the original model and replacement method, on two datasets are shown in Table~\ref{tab:results}. It can be found that our method outperforms them under most conditions in both n-gram accuracy and diversity evaluation. Compared to the original models, our method (EA) can significantly promote the n-gram diversity of responses and slightly benefit n-gram accuracy under most conditions while remaining equivalent under other cases. In other words, EA encourages models to generate diverse texts without the cost of their quality. In contrast, the replacement method usually degrades the generation accuracy despite it is also beneficial for more various replies. 

When comparing between base models, EA seems to have more effect on transformer than Seq2seq. The reason is that more data is needed to train a robust transformer than RNN due to its depth, and our method pose a similar effect. It can also be observed that the PPL values of both EA and replacement can result in a noticeable PPL decrease for transformer models, but it is not such a case for Seq2seq models. It means transformers may not be well trained using current raw data. 

We also evaluate the performance of Seq2seq / transformer-BERT to better prove the merits of using fastText as a semantic-neighbor word prediction model. The performance of them is shown in Table~\ref{tab:bert} in terms of accuracy metrics, Ent, Dist, and Sen(the average of Ent/Dist/Sen-n and n=1, 2, 3). EA and BERT variant show equivalent performance on both base models, because BERT embeddings without tuning on a specific domain can even perform worse than a universal embedding~\cite{bert_embedding}. However, the huge BERT model will increase the computational load since the training time of our EA is approximately 1.98$\times$ time of raw models, while it is 5.87$\times$ for BERT variant and 1.37$\times$ for replacement according to experiments. Obviously, it is better to select fastText rather than BERT for augmentation candidates prediction.

\subsection{Ablation study}

We also conduct the ablation study including the following variants: 1) w/o soft label: we only use the original hard labels during training which is the same as the previous models; 2) w/o history augmentation(aug.): we only applying embedding augmentation to response $R$. The related results of two models on DailyDialog dataset is shown in Table~\ref{tab:ablation}. We easily find out the contribution of both parts according to the performance degeneration under most conditions.

\begin{table}[t!]
\setlength{\tabcolsep}{0.9mm}
\small
\centering
\begin{tabular}{l|c|cc|ccc}
\hline 
\textbf{Model} & PPL & BLEU & N-4 & Ent & Dist & Sen  \\
\hline
Seq2seq-EA & 53.541 & 1.214 & 0.217 & 6.253 & 19.699 & 95.057 \\
\quad w/o soft label & 55.454 & 0.861 & 0.098 & 6.065 & 19.160 & 94.597 \\
\quad w/o history aug. & 59.684 & 0.956 & 0.134 &  6.024  & 17.606 & 94.105 \\
\hline
transfo-EA & 74.020 & 1.097 & 0.198 & 5.349 & 9.637 & 94.692 \\
\quad w/o soft label & 72.135 & 0.879 & 0.100 & 5.241 & 9.182 & 94.127 \\
\quad w/o history aug. & 77.499 & 0.985 & 0.139 & 5.221 & 9.536 & 94.381 \\
\hline
\end{tabular}
\caption{\label{tab:ablation}Results of the ablation study on DailyDialog dataset.}
\end{table}

\begin{figure}[ht!]
\setlength{\belowcaptionskip}{1cm}
\centering
\includegraphics[width=0.95\linewidth]{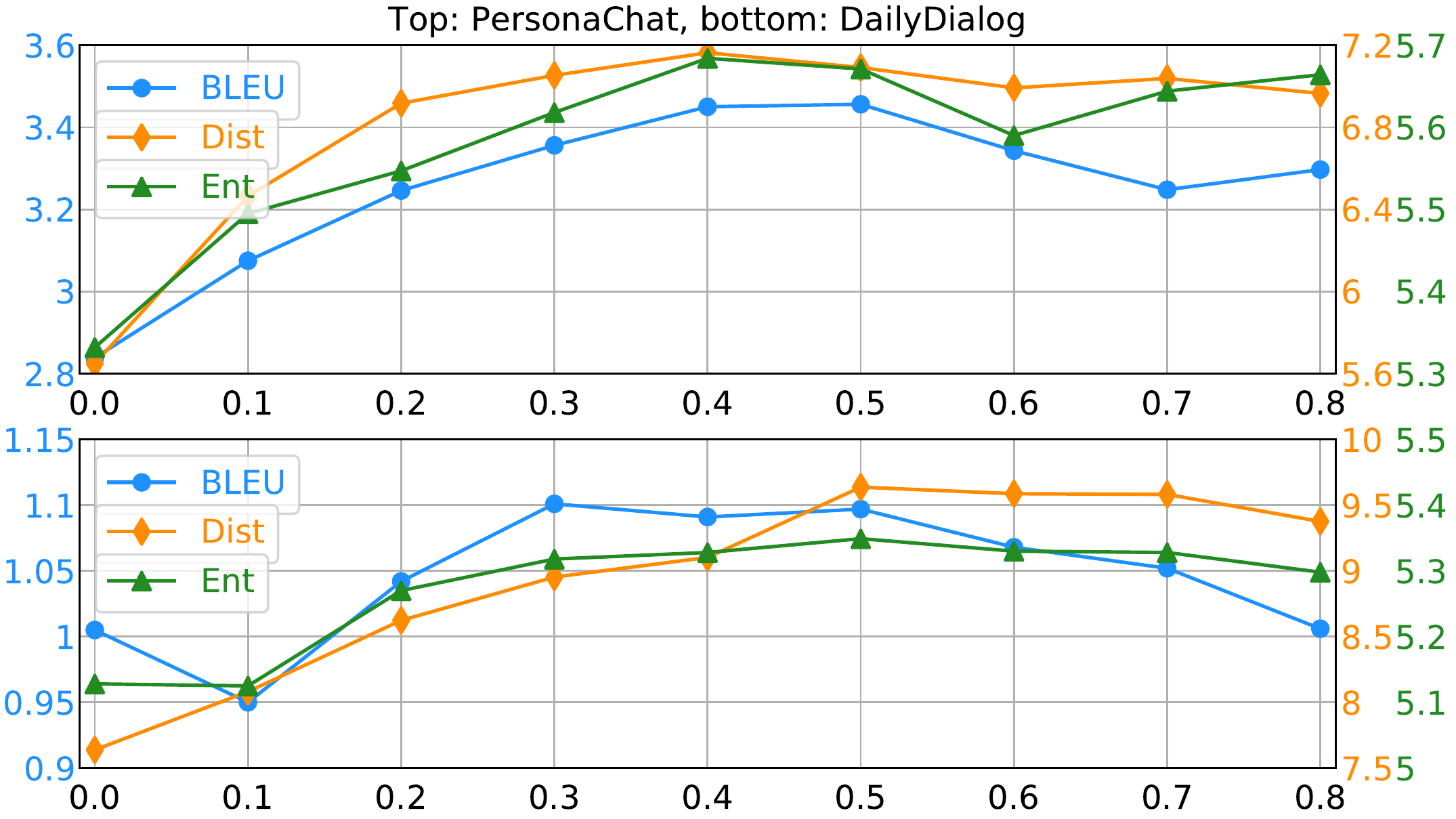}
\caption{The performance of n-gram accuracy and diversity of transformer-EA varies with different augmentation ratio $\rho$.}
\label{fig:ratio}
\end{figure}

\subsection{The influence of augmentation ratio}

To better illustrate the influence of different augmentation ratio $\rho$, we also tried different $\rho$ values from 0 to 0.8. The BLEU score (for n-gram accuracy), Dist, and Sen values (for diversity) of transformer model on two datasets vary with $\rho$ is shown in Fig.~\ref{fig:ratio}. In the beginning, both BLEU score and diversity metrics tend to increase slowly. But after a middle value (0.4 on PersonaChat and 0.5 on DailyDialog), BLEU scores begin to drop while generation diversity is also becoming saturated. A higher ratio than these peak values will not benefit the performance but merely increase the training time as more neighbor prediction and embedding fusion operations are needed. That is the reason why we choose these $\rho$ values.

\section{Conclusion}

Considering the many-to-many essence of the dialogue generation problem, we propose an efficient embedding augmentation method aiming to promote the response diversity. Different from previous work, our manipulation is based on embedding level instead of token level by transforming a soft word, which is a distribution between the original word and its semantic neighbors, into a fused embedding to randomly replace the original one. We also use such a distribution as a soft label to realize our multi-target training goal. The experimental results demonstrate that our method can effectively diversify the produced replies from both raw Seq2seq and transformer models with even slight improvement in accuracy, and it is also superior to replacement-based methods.
\fontsize{10pt}{11pt} 
\selectfont
\bibliographystyle{ICASSP2021}
\bibliography{ICASSP2021.bib}

\end{document}